# Geospatial AI, Dataverse Metadata, and the Study of Place-Based Government

Danny EBanks[1] and Devika Jain[1]

[1] Harvard University

## 1. Introduction

Harvard Dataverse is the largest open repository for social science research data in the world, hosting over 150,000 datasets deposited by researchers across every empirical discipline.[1]

Founded at Harvard's Institute for Quantitative Social Science as an infrastructure for publishing, preserving, citing, and reusing research data (King 2007; Crosas 2011), the repository has become a central node in the ecosystem of open science, with native support for geospatial, social science, life science, and biomedical data and their field-specific metadata schemas. We, in turn, show that graph-theoretic methods can transform this collection of open-source data into the foundations of a searchable infrastructure for geospatial research, and we demonstrate how in this chapter. To achieve this, we construct a knowledge graph using the data and metadata made publicly available on the Harvard Dataverse repository, using it as a framework organizing 102,650 datasets inside a 215,985-node network with 528,003 edges, and 43,991 of those datasets carry geospatial metadata that records where the data apply, the spatial resolution of the observations, and the geographic boundaries of the study area. This network provides a foundational framework and means of analyzing geospatial data as a problem of place resolution, entity linking, and cross-dataset navigation. The contribution of this chapter is both empirical and methodological in nature; for example, when we rely on keyword search and search over titles, descriptions, and keywords for a short list of policy-related terms, 7,654 geospatial datasets prove to be directly relevant to government, legislation, health policy, education policy, transport, planning, or policing. Seven thousand datasets, organized inside a quarter-million-node network, are enough to reveal how geospatial research is structured across disciplines, but the metadata behind those datasets is unstructured enough that place resolution remains a persistent problem, and the network topology we analyze in this chapter provides the structural framework for making that problem tractable.

We proceed with this chapter as follows. Section 2 explains the distinguishing features of a research data repository as opposed to a file storage software and where geospatial metadata fits within the larger data repository ecosystem. Section 3 reports the extent of the geospatial metadata block. We then show, in Section 4, that a surprising share of the geospatial metadata corresponds with datasets directly related to elections, legislatures, public health, education, and municipal government. Section 5 walks through specific datasets and their applications. Section 6 develops an extended use case that adds language to place and maps the geography of partisan discourse and political conflict across American communities, and Section 7 discusses limitations and future avenues of research.

## 2. Research Data Repositories and the Role of Geospatial Metadata

The distinction between a repository and a file server matters here because it determines the role that metadata can play. File servers store and retrieve bits, and they make no promises beyond that narrow function. Repositories do something categorically different; they manage persistent identifiers, version histories, access controls, structured metadata fields, and links to publications, all so that a dataset remains discoverable and interpretable long after the original investigators have left the project. Harvard Dataverse is one such repository, and it serves as the empirical setting for the knowledge graph we construct and analyze in this chapter. Every one of those repository functions depends on metadata. Without a structured description, a dataset is findable only by someone who already knows it exists (which undermines the purpose of depositing it in a shared archive), while rich metadata allows the same dataset to surface in a search, link to neighboring records, and participate in machine-readable queries that the depositor never imagined when uploading the files.

A knowledge graph provides a natural representational framework for this kind of structured metadata. We use the term in the sense established by Hogan et al. (2021) and Ehrlinger and Wöß (2016); a knowledge graph is a labeled, directed network whose nodes represent entities (in our case, datasets, keywords, publications, geographic locations, subjects, and journals) and whose edges represent typed relations among those entities (has keyword, has location, has subject, related publication, and so on).[2]

Knowledge graphs have become a standard tool for organizing heterogeneous information in settings where the relationships between entities are as important as the entities themselves; Google's Knowledge Graph, introduced in 2012, demonstrated at commercial scale that graph-structured metadata could power semantic search, and the approach has since been adopted for biomedical data (Himmelstein et al. 2017), scientific literature (Ammar et al. 2018), and open research knowledge graphs (Auer et al. 2018). The property that makes knowledge graphs particularly well-suited to a research data repository is that they preserve the typed, heterogeneous structure of the metadata rather than collapsing it into a flat index. A dataset in the Knowledge Graph on Harvard Dataverse is connected to its keywords by one type of edge, to its geographic location by another, and to the publications that cite it by a third, and each of those edge types carries different information about the dataset's content, geographic scope, and relationship to other work in the repository. Figure 1 illustrates this structure for a small neighborhood of the graph, showing how four policy-relevant datasets are linked through shared keywords, subjects, journals, and geographic locations.

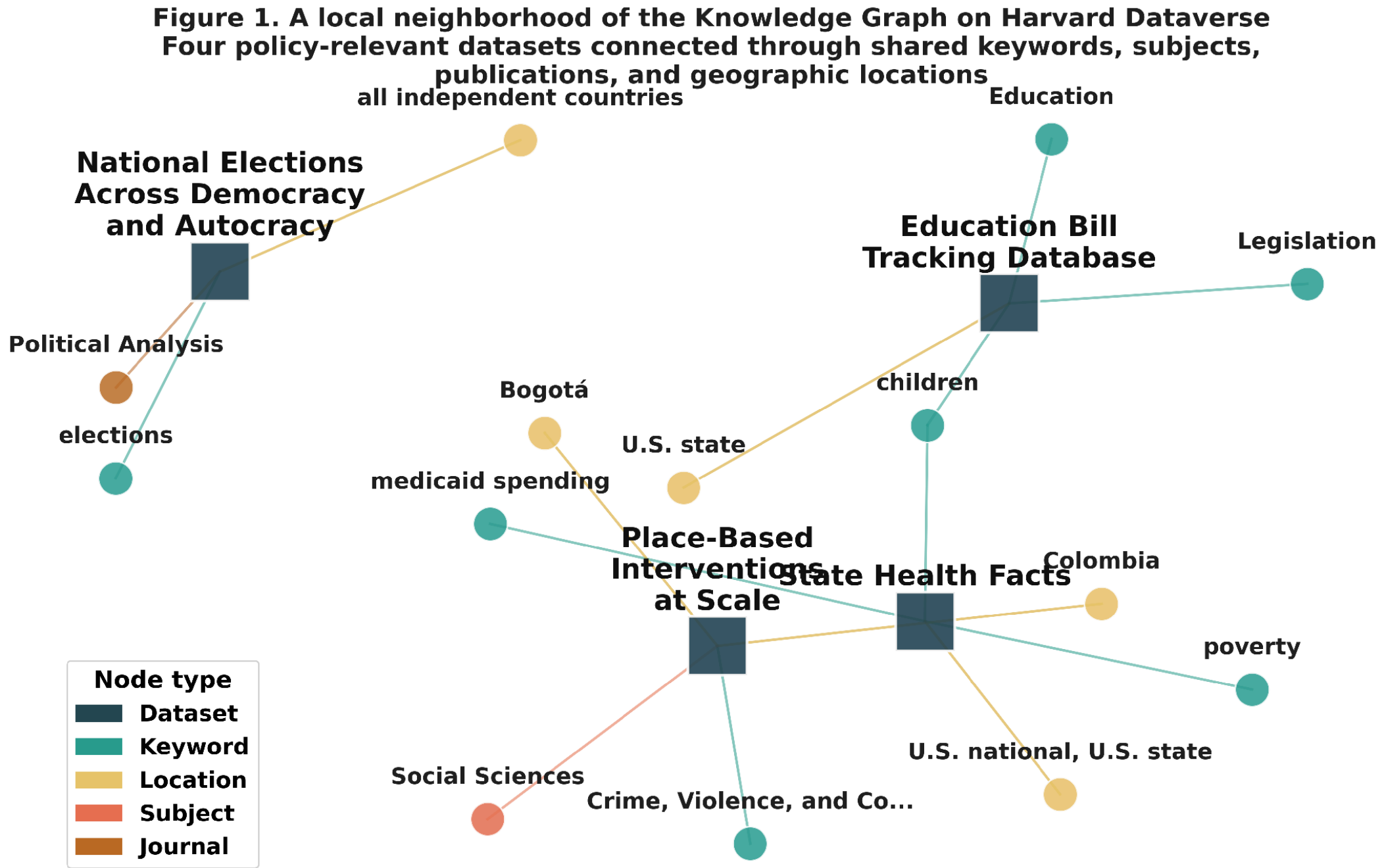

*Figure 1. A local neighborhood of the Knowledge Graph on Harvard Dataverse, showing four policy-relevant datasets connected through shared keywords, subjects, journals, and geographic locations.*

Geospatial metadata fills a narrower but well-defined niche inside this infrastructure. It records three things about a dataset's geographic scope; (1) where the dataset applies (a named country, state, city, or region), (2) the spatial unit at which observations are measured (county, census tract, polling station, grid cell), and (3) when coordinates or bounding boxes are available, the geographic envelope that bounds the dataset's footprint on a map. The 43,991 geospatially tagged datasets in the Knowledge Graph on Harvard Dataverse each carry at least one of these fields, and that overlap is the feature that makes location a searchable, linkable, and comparable property of each repository record. A researcher looking for county-level health data in the southeastern United States can, in theory, find every dataset that meets those criteria; the difficulty is that the metadata has to be structured and the place names have to resolve to a common reference frame.

The difficulty in finding high-quality geospatial metadata is largely owed to the fact that raw geospatial metadata in Dataverse is provided by data depositors in free text form with no controlled vocabulary to constrain the choice set facing the depositor, which means a dataset tagged "New York" might refer to the state, the city, or a single borough, and the repository has no machinery to disambiguate the difference. Even if providers furnish a bounding box, there is no guarantee this will resolve the issue. For example, if the geographic borders provided by the depositor cover the continental United States, such information tells an analyst little that is meaningful about whether the underlying data describe fifty states or a single county in Ohio. Noting these ambiguities is not to diminish the value of the data that depositors have already provided; they highlight and make concrete the missing data problem that geospatial AI aims to alleviate. In essence, that problem is to take the metadata depositors provide, resolve it to some canonical place, and integrate it with the larger structure (in our case, a knowledge graph) of datasets, keywords, and publications that the repository already maintains.

## 3. The Geospatial Metadata Layer

Nearly all of the Knowledge Graph on Harvard Dataverse sits in one giant connected component, and that single structural fact determines the kinds of relationships the graph can and cannot reveal about the organization of geospatial research. We constructed the graph from 215,985 nodes and 528,003 edges; of those nodes, 102,650 are datasets, 56,956 are keywords, and 48,198 are publications, with 96.9 percent belonging to that dominant component, which means that most datasets are reachable from most other datasets through short chains of shared keywords, shared authors, or co-citation with the same publications, so a query that starts from one geospatial dataset can traverse the graph to find neighbors even when the geospatial metadata fields between the two records share nothing in common.

Of the 102,650 dataset nodes, 43,991 (42.9 percent) carry geospatial metadata, and three fields define this geospatial layer of the graph. Geographic coverage, which is simply a text string naming the place or places the data describe, appears for 43,736 datasets and is by far the most common. Geographic unit, which records spatial resolution (city, county, state, country, grid cell), appears for 30,798. Bounding boxes, which encode latitude-longitude coordinates defining the geographic envelope, appear for 28,492. These fields overlap considerably, and the ordering by frequency tracks how much work each field demands of the depositor; typing "United States" into a text field takes seconds, while constructing a proper bounding box requires coordinates that many depositors never computed for their own analyses and have no reason to retrieve for a metadata form.

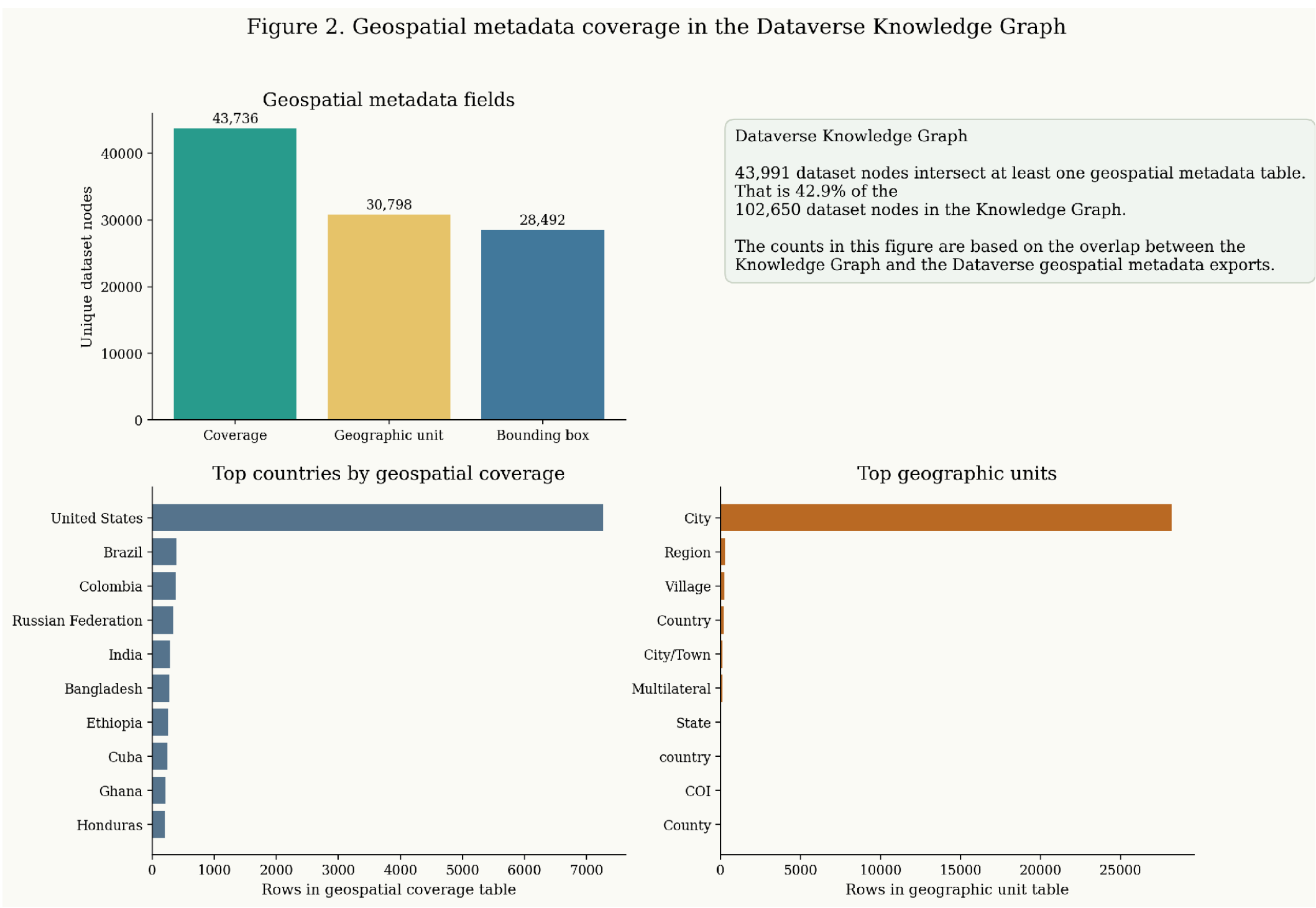

*Figure 2. Geospatial metadata coverage in the Knowledge Graph on Harvard Dataverse.*

City is the dominant geographic unit in the knowledge graph, reflecting which disciplines deposit there most frequently, and the United States leads all countries in geospatial coverage records because American researchers predominate among depositors (Figure 2 shows the distribution along both of these dimensions). The concentration is a feature of who uses the repository rather than a design choice, but it constrains the analyses the current metadata can support in practice; analysis built on American municipal and state data will be well grounded, while extending it to countries where depositors supply coverage labels without coordinates will require supplementary geocoding, and we return to this constraint in Section 7.

**4. Government and Local Policy in the Geospatial Metadata**

Geospatial metadata in a research repository brings to mind environmental science, remote sensing, and satellite imagery, and for a large fraction of the repository that association holds. We searched titles, descriptions, and keywords across the 43,991 geospatial datasets for terms related to government, legislation, health policy, education policy, transportation, urban planning, and policing, and that search identified 7,654 datasets (17.4 percent) as directly policy-relevant. A broader set of search terms would capture more, so 7,654 is a conservative lower bound. Even at that lower bound, the finding reveals that political scientists, public health researchers, and policy analysts have been populating the geospatial metadata fields in Dataverse for years (perhaps without anyone planning it this way) simply because their research questions demanded geographic identifiers.

In terms of the research topics uncovered in the network structure, elections and legislatures account for the largest cluster, comprising 4,286 datasets, or 56 percent of the policy-relevant geospatial total. The next largest cluster, government and administration, follows with 1,719 datasets, health policy with 1,115, transportation and transit with 1,019, and education policy with 966; these categories are not mutually exclusive, so the statistics do not sum to 7,654 (Figure 3 shows the full distribution). The dominance of the electoral category reflects decades of data-sharing norms in American electoral research; precinct-level returns, campaign finance records, and legislative roll calls have been deposited with state and district identifiers since well before the repository adopted structured geospatial fields. In a connected, but certainly distinct domain of research, we find there are 1,115 health policy datasets, a cluster large enough to constitute its own subgraph and one whose geospatial tagging follows a different logic than the electoral data.

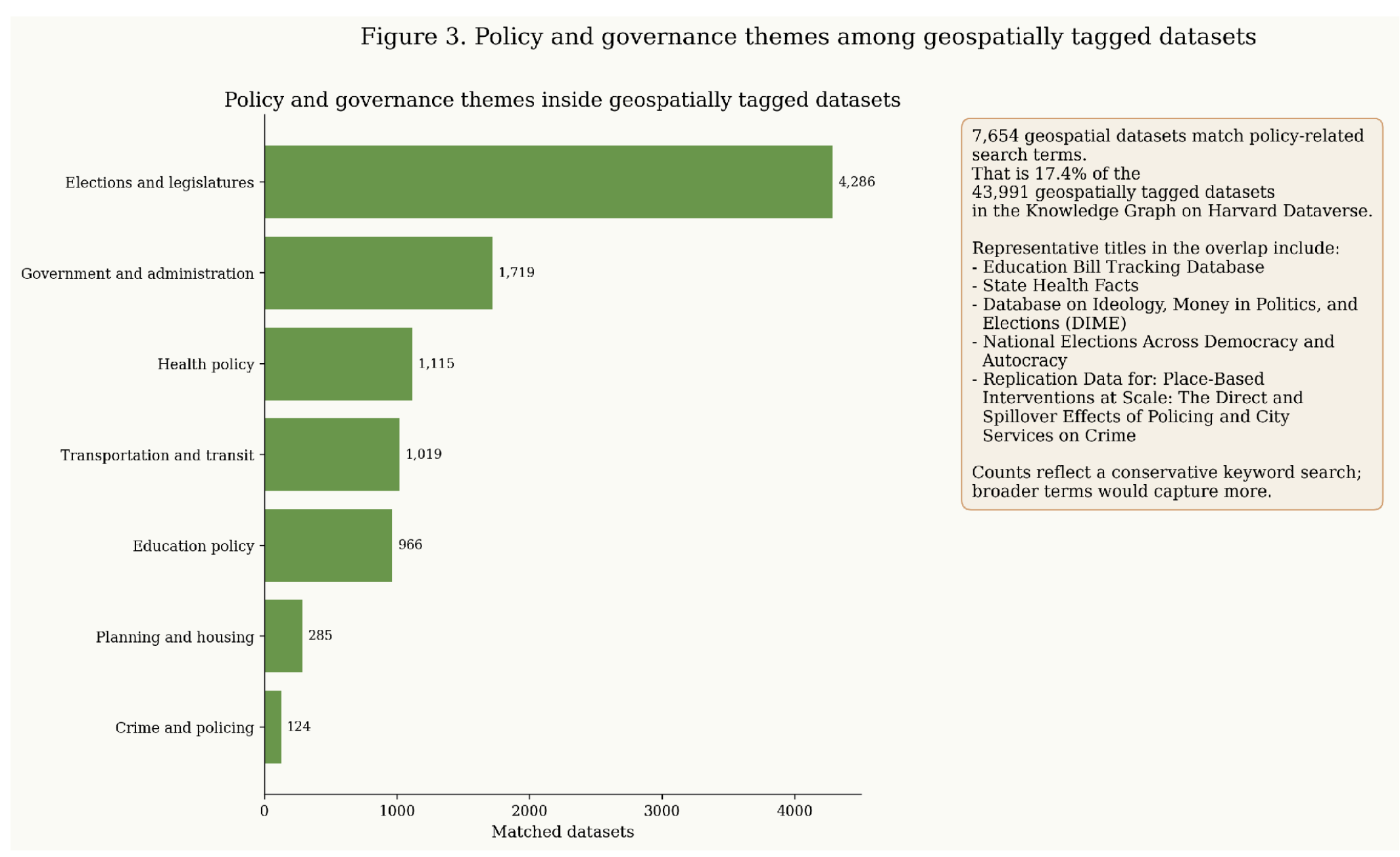

*Figure 3. Policy themes among geospatially tagged datasets.*

More so than other fields, the domain of public health research is spatial by construction; disease incidence varies by county, hospital access varies by zip code, environmental exposures vary by census tract, and none of those spatial dimensions are optional in the research design. Researchers who deposit public health data in Dataverse tend to fill in the geographic coverage and geographic unit fields because the spatial structure of their data is integral to the study design itself. Transportation researchers deposit route maps, station coordinates, and service-area boundaries that are geographic by definition, and education researchers deposit school-district boundaries, attendance zones, and state testing data that carry place-based structure into the metadata whether or not the depositor thinks of the data as "geospatial." We conclude from this pattern that geospatial metadata in Dataverse indexes a body of policy-relevant evidence spanning county-level health outcomes, state-level education legislation, municipal policing experiments, and federal transportation networks.

## 5. Empirical Examples

To better understand how the network structure better enables us to understand the geospatial component of the larger universe of research repository data, we turn to five individual datasets that show how that metadata works in practice across different policy domains, spatial scales, and research designs. The Education Bill Tracking Database records state-level education legislation session by session, with geographic coverage tagged at the state level and legislative chamber as the unit of analysis. State Health Facts operates at a finer grain; it collects county- and state-level indicators for health outcomes, insurance coverage, and provider supply, and its bounding box metadata is precise enough to support spatial queries down to the county level (a rarity in the repository, as we noted in Section 3).

The Database on Ideology, Money in Politics, and Elections (DIME) links campaign finance records to candidates and donors at the district, state, and national level, with geographic identifiers that permit place-based analysis of political spending across all three scales simultaneously. National Elections Across Democracy and Autocracy compiles election returns from dozens of countries and decades of coverage, tagged with country-level geographic metadata. Place-Based Interventions at Scale, by contrast, works at the municipal level, recording experimental and quasi-experimental evidence on policing and city services with city-level geographic metadata that pins each study to a specific jurisdiction.

These five datasets were deposited by different, individual researchers, in different years, for entirely different purposes. Yet, geography connects them despite these different initial purposes; each carries structured geospatial metadata, and the knowledge graph links them through shared location nodes even when they share no keywords or publications. That convergence (independent research projects finding each other through place rather than through citation) is the empirical foundation of this chapter. A researcher studying how policing interventions affect educational outcomes in the same city could follow the knowledge graph from Place-Based Interventions at Scale to the Education Bill Tracking Database through their shared state-level geographic tags, then outward to related publications and co-occurring keywords. We face a problem of representation and resolution; we need to link records by place so that the network structure can surface connections that no keyword search would find on its own.

## 6. Mapping the Language of Conflict with Generative AI

The framework developed so far links datasets by place, but it says nothing about the discourse those communities produce. We close that gap with an extended use case that measures partisan conflict in language and locates it across American communities.

Political polarization is not merely an attitudinal phenomenon; it is a linguistic one. Gentzkow, Shapiro, and Taddy (2019) analyzed 13.5 million documents comprising 1.8 billion words spoken by 37,059 senators and representatives in the U.S. Congressional Record between 1873 and 2016, and their estimates show that partisanship in congressional speech increased sharply in the early 1990s and has remained at historically elevated levels since. The same divergence operates at the mass public level. Conover et al. (2011) examined more than 250,000 tweets from the six weeks leading up to the 2010 U.S. congressional midterm elections and found that the network of political retweets exhibited a highly segregated partisan structure, with extremely limited connectivity between left- and right-leaning users; the mention network, by contrast, was dominated by a single politically heterogeneous cluster, suggesting that users inject partisan content precisely into information streams whose primary audience consists of ideologically opposed individuals. Bail et al. (2018) extended this line of work with a field experiment on Twitter, offering Democrats and Republicans financial compensation to follow bots that retweeted messages from elected officials and opinion leaders with opposing political views; Republicans who followed a liberal bot became substantially more conservative post-treatment, a finding that complicates the intuition that exposure to opposing views reduces polarization. Vosoughi, Roy, and Aral (2018) investigated the differential diffusion of approximately 126,000 verified true and false news stories distributed on Twitter from 2006 to 2017 and found that falsehood diffused significantly farther, faster, deeper, and more broadly than the truth in all categories of information, with the effects most pronounced for false political news.

These findings from political science and computational social science establish three facts that the GeoAI-for-Policy framework must accommodate. First, partisan divergence is encoded in language itself, not only in voting behavior or survey responses, which means that text data are a direct observable for measuring polarization. Second, the network structure of online discourse amplifies partisan segregation through selective sharing and algorithmic curation, so that the *topology* of who talks to whom matters as much as the *content* of what they say. Third, the geography of online discourse is not uniform; participation, attention, and stance vary by state, by metropolitan area, and by neighborhood, which means that any framework linking text to policy must account for spatial heterogeneity. We proceed in this section to describe three methodological contributions—community language models, stance detection pipelines with geospatial aggregation, and partisan language bridging tools—that operationalize these facts for the kinds of place-based policy analysis the chapter envisions.

### 6.1 Probing Partisan Worldviews with Community Language Models

One of the most direct ways to measure ideological conflict in discourse is to build separate language models for distinct political communities and then compare what those models generate when prompted with identical questions. The CommunityLM framework (Jiang, Beeferman, Roy, and Roy 2022) operationalizes this idea by constructing a Twitter dataset containing 4.7 million tweets (100 million word tokens) from Republican and Democratic communities, respectively. The study first sampled 1 million active U.S. Twitter users from the Twitter firehose (a 10 percent sample of all tweets) who posted at least 10 original tweets before and after the 2020 presidential election, used Litecoder to extract user locations from profile strings and filter out non-U.S. users, and then constructed the follow graph of the resulting 1,074,650 Twitter users. Partisan assignment followed the method of Volkova et al. (2014) and Demszky et al. (2019): a user was labeled as a Democrat if they followed no fewer than 6 Democratic politicians and no Republican politician from a list of 457 Republican and 473 Democratic politician handles, and as a Republican if they followed no fewer than 2 Republican politicians and no Democratic politician. This procedure identified 182,788 Democratic-leaning and 72,186 Republican-leaning users (a ratio of 2.53:1, consistent with Pew Research Center estimates of the partisan composition of Twitter).

The framework then fine-tuned GPT-2 language models on the tweets authored by each community and assessed the worldviews of the two groups using prompt-based probing with prompts that elicit opinions about 16 public figures and 14 groups surveyed by the American National Election Studies (ANES) 2020 Exploratory Testing Survey, which was conducted between April 10 and April 18, 2020, on 3,080 adult citizens from across the United States. Fine-tuned CommunityLM with the prompt format “X is/are the” achieved 97.33 percent accuracy and 97.29 percent weighted F1-score on the task of predicting which community is more favorable towards each of the 30 survey items, outperforming all baselines including pre-trained GPT-3 Curie (84.02 weighted F1) and keyword retrieval methods (93.33 weighted F1). Figure 4 shows the resulting rankings of 16 public figures by their average stance scores from the fine-tuned Democratic and Republican CommunityLM models; Republican politicians are rated poorly by the Democratic model and vice versa, and the ratings from the Republican model are more negative overall than those from the Democratic model.

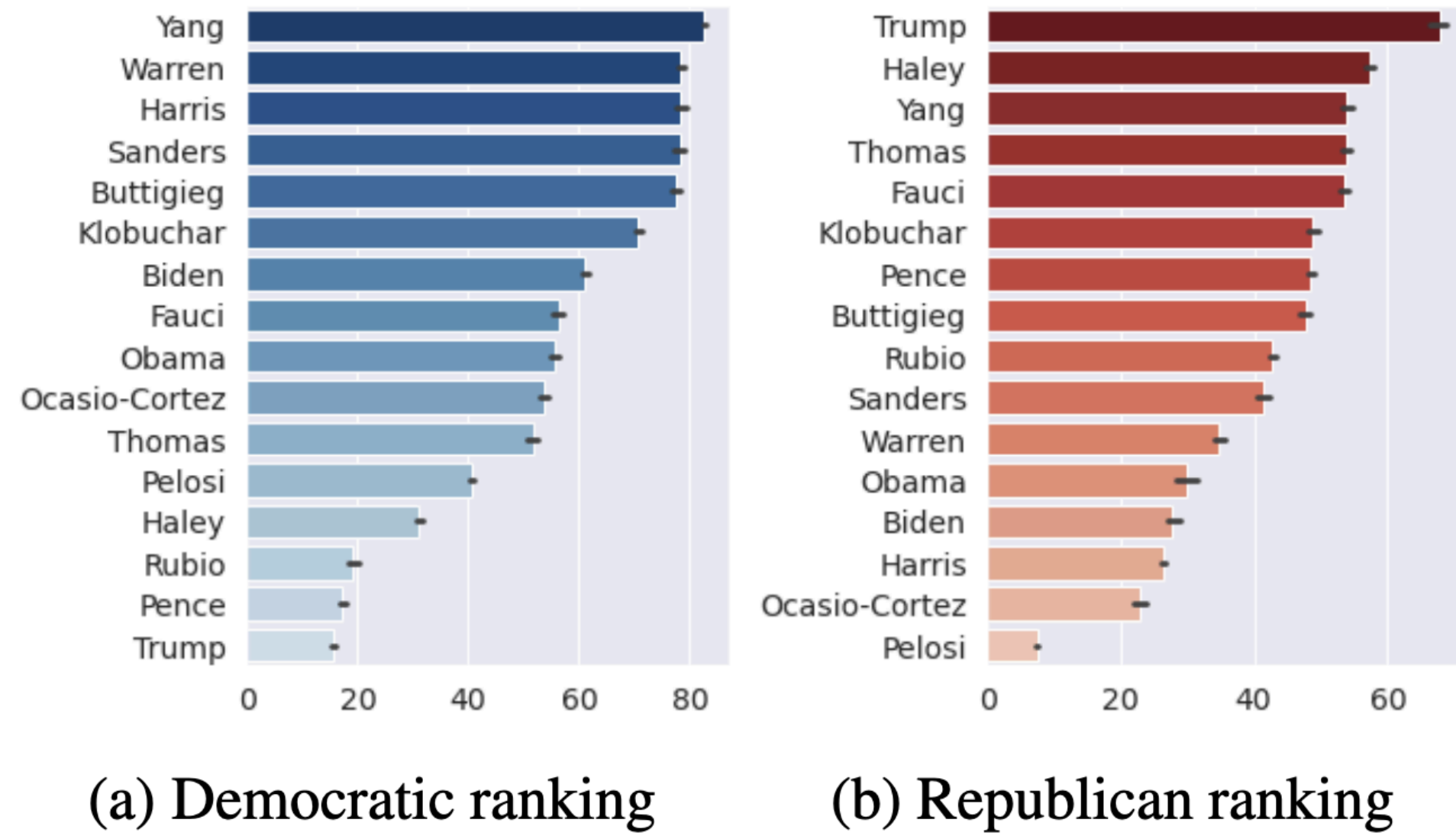

*Figure 4. Left and right rankings of 16 public figures by their average stance scores calculated on synthetic tweets from fine-tuned CommunityLM models. Source: Jiang et al. (2022).*

The key methodological insight is that language models trained on community-specific text internalize the statistical regularities of that community's discourse, including its evaluative stances, its favored framings, and its patterns of attention and omission. This finding builds on earlier work in computational political ideology detection; Iyyer et al. (2014) demonstrated that recursive neural networks could identify the political position expressed by a sentence at the sub-sentential level using crowdsourced annotations, while Preoțiuc-Pietro et al. (2017) moved beyond binary liberal-conservative labels to predict political ideology on a seven-point scale from Twitter data. CommunityLM extends these approaches by using generative language models rather than classifiers, which allows not just detection of ideology but *probing* of community worldviews on arbitrary topics.

For the GeoAI-for-Policy framework, CommunityLM offers a scalable instrument for measuring one of the chapter's central quantities: the spatial distribution of partisan worldviews. Because social media posts carry geographic metadata—geotagged coordinates, user-reported locations, or inferred locations from network and content features—community language models can in principle be trained not only for national-level partisan groups but for geographically defined communities, that is, users in specific states, metropolitan areas, or congressional districts. Barberá (2015) demonstrated that the follow-graph structure of Twitter encodes ideological positions that can be estimated via Bayesian ideal point models for millions of users across the United States and five European countries; coupling that kind of geographic ideological estimation with community-specific language model training would produce place-based indicators of worldview divergence at whatever spatial resolution the analyst requires.

**6.2 Stance Detection and the Geography of Public Health Discourse**

While CommunityLM captures the structure of partisan worldviews at the community level, policy-relevant conflict often crystallizes around specific contested topics where stance—support, opposition, or neutrality towards a claim or intervention—varies across geography, demography, and time. The study of public perceptions of COVID-19-related medications (Hua, Jiang et al. 2022) demonstrates how NLP pipelines can be deployed to track the geographic distribution of conflictual discourse in real time during a public health crisis, and the pipeline architecture it introduces generalizes to any policy domain where public stance matters.

The study analyzed 609,189 U.S.-based tweets spanning January 29, 2020, through November 30, 2021—93 weeks of data divided into three sections based on the onset of the three major waves of COVID-19 in the United States—focusing on four drugs that generated significant public attention: hydroxychloroquine and ivermectin (therapies promoted with anecdotal evidence but lacking robust clinical support for COVID-19) and remdesivir and molnupiravir (FDA-authorized treatments for eligible patients). The NLP pipeline, shown in Figure 5, combined several components: two COVID-drug-stance-BERT models fine-tuned from a pretrained Twitter-specific RoBERTa-base stance detection model on the COVID-CQ stance dataset (with 14,353, 14,373, and 14,372 tweets labeled as negative, neutral, and positive, respectively), achieving 86.88 percent accuracy on validation data; BERTweet embeddings with K-Means clustering (k = 15) for content analysis to identify thematic arguments; and named entity recognition using Stanza's 4-class tweet NER model (person, organization, location, and miscellaneous) to extract influencers and organizations from discourse across each pandemic wave.

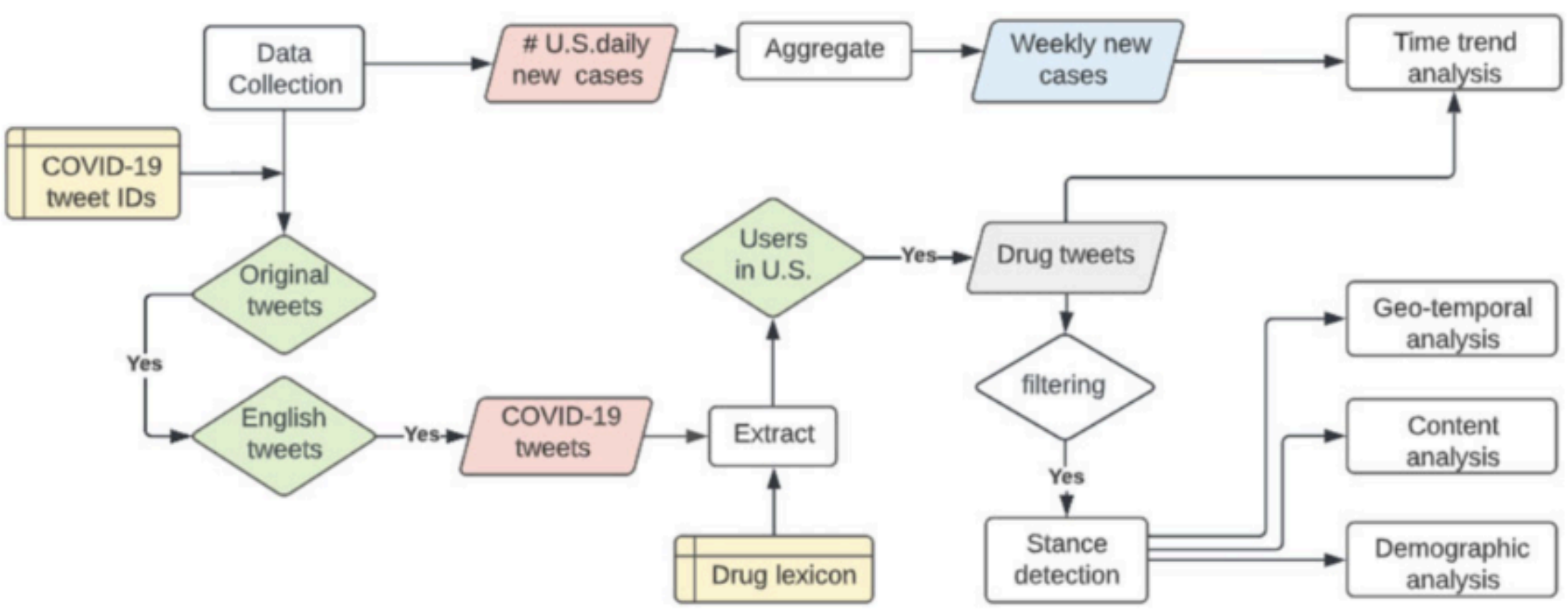

*Figure 5. A comprehensive multimodal pipeline to study the public perception of drugs during the COVID-19 period. Source: Hua, Jiang et al. (2022).*

The time-trend analysis revealed that off-label drugs dominated public discussion: hydroxychloroquine and ivermectin accounted for 255,573 (42.0 percent) and 332,381 (54.6 percent) of drug-related tweets respectively, while molnupiravir and remdesivir received sporadic attention (6,285 and 54,950 mentions, or 1.0 percent and 9.0 percent). Most tweets showed a supportive attitude (53.04 percent and 57.72 percent for hydroxychloroquine and ivermectin, respectively), and Figure 6 shows both the temporal trajectories of drug-related tweets alongside weekly COVID-19 case counts and the distribution of stance across the three pandemic waves.

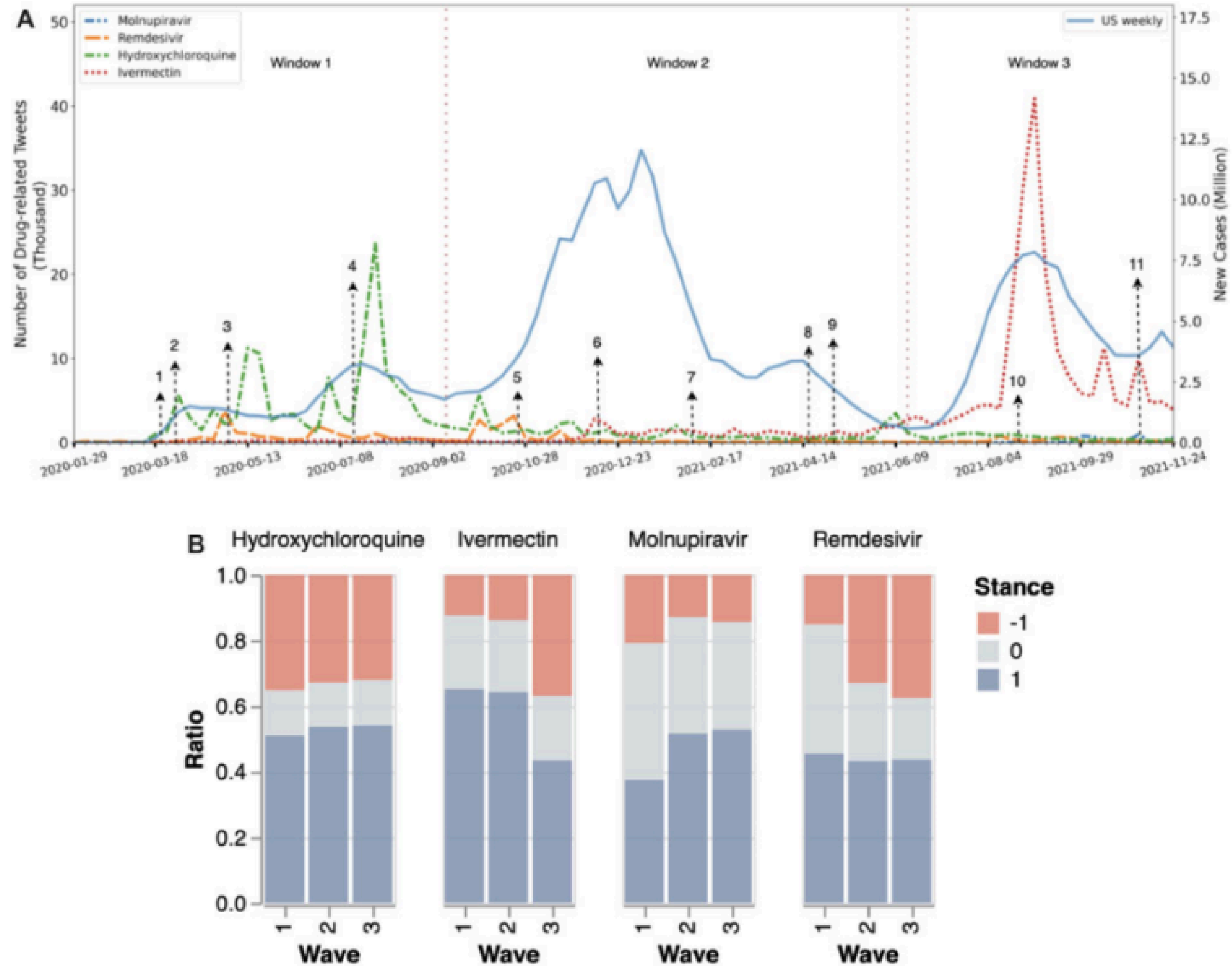

*Figure 6. (A) Time trends of tweets mentioning COVID-19-related drugs alongside weekly U.S. COVID-19 case counts (stepped line). (B) Distribution of positive (blue), neutral (gray), and negative (red) stances for each drug across the three pandemic waves. Source: Hua, Jiang et al. (2022).*

The geographic and demographic dimensions of the conflict were stark. The political partisanship assignment method (following Demszky et al. 2019 to infer political affiliation) predicted 45.5 percent Democratic, 16.3 percent Neutral, and 38.3 percent Republican from the total sample, and the distribution of stance for left and right partisans was significantly different (Chi-square test) for all four drugs ($P < .001$ for all). Republicans were much more likely to support hydroxychloroquine (71 percent vs. 16 percent) and ivermectin (63 percent vs. 33 percent) than Democrats. People with healthcare backgrounds tended to oppose hydroxychloroquine (48.0 percent vs. 41.7 percent of the general population) and support molnupiravir (49.3 percent vs. 42.0 percent).

The longitudinal geo-temporal analysis, shown in Figure 7, visualized the average statewide stance per drug per wave for geoinference, revealing that these patterns were not merely reflections of national partisan identity but exhibited regional variation. Hydroxychloroquine had an average positive sentiment manifested in Wyoming and Alabama in waves 1 and 2, and Louisiana in wave 3. Ivermectin had positive attitudes primarily in Montana, Arizona, Florida, and other states, but only in waves 1 and 2. Molnupiravir had consistent positive discussions in most states that had data, while Remdesivir had mostly neutral discussions except in wave 3.

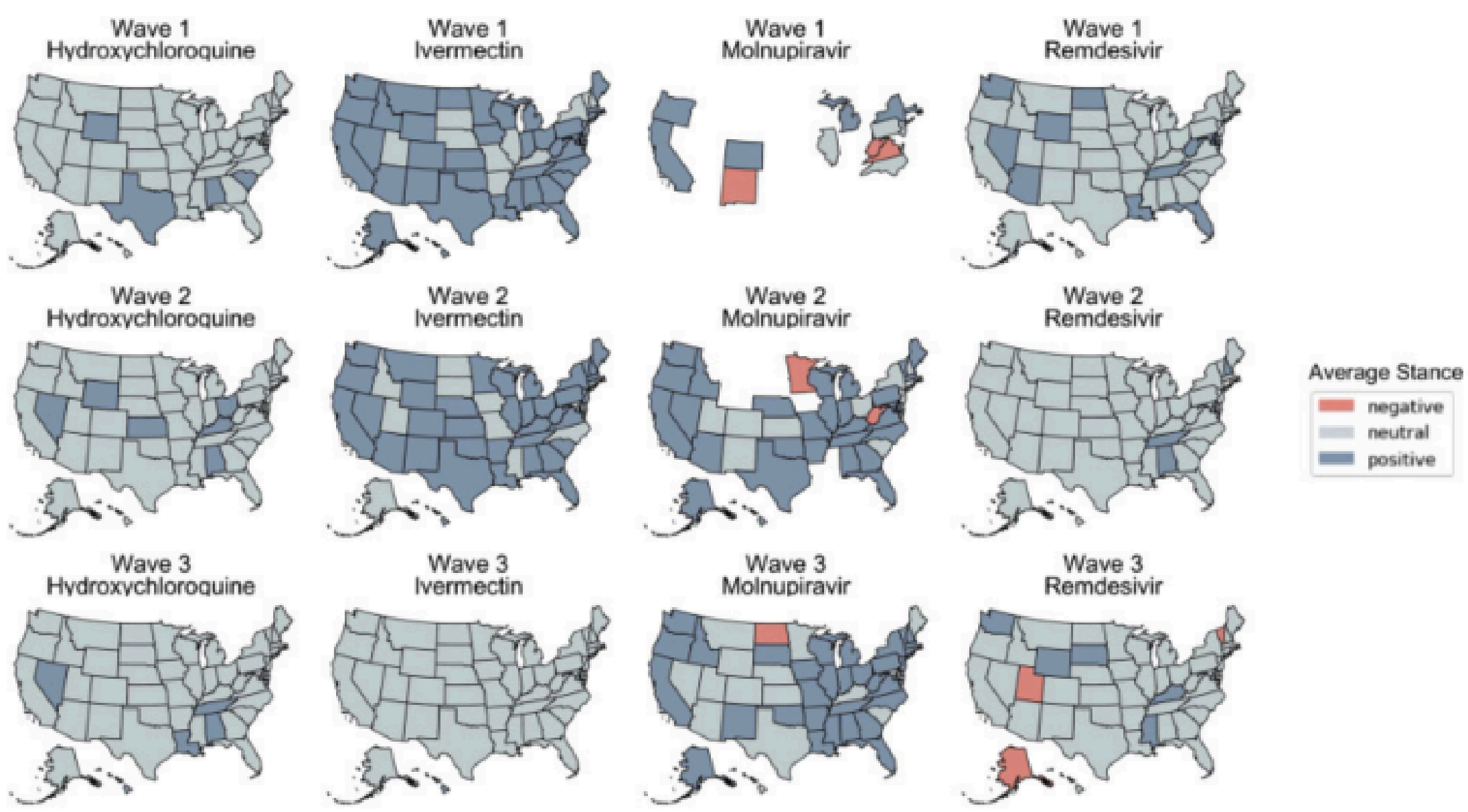

*Figure 7. Longitudinal geo-temporal analysis of tweeted sentiment of the four drugs by COVID-19 pandemic wave. The average sentiment of each state was classified into positive, neutral, and negative. Source: Hua, Jiang et al. (2022).*

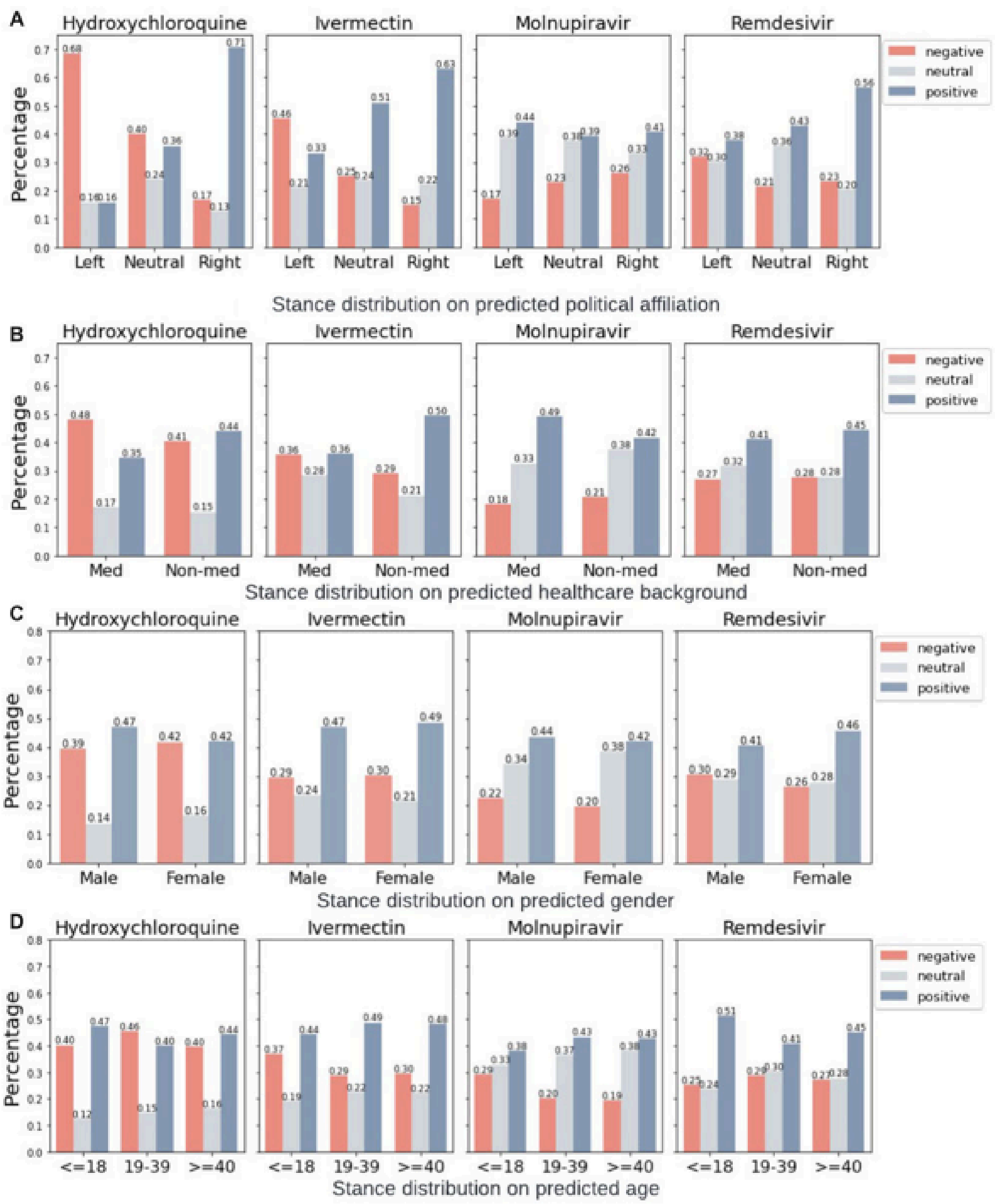

*Figure 8. Stance distribution on predicted partisanship, healthcare background, gender, and age for each drug. Source: Hua, Jiang et al. (2022).*

This study illustrates several principles central to the chapter's framework. First, the combination of stance detection with geographic aggregation produces precisely the kind of place-based indicators the opening section envisions: maps of where conflictual discourse concentrates, how it evolves over time, and which communities amplify or resist specific narratives. Second, the finding that celebrity and political figures—particularly Donald Trump for hydroxychloroquine and Joe Rogan for ivermectin—significantly amplified contested claims demonstrates how diffusion pathways interact with geography; a tweet from a national political figure does not land uniformly across the country, and its amplification depends on the density of sympathetic audiences in specific locations. This finding aligns with Vosoughi, Roy, and Aral's (2018) observation that false news travels faster and farther than true news, and extends it by showing that the geography of misinformation susceptibility is itself patterned by partisan and demographic composition. Third, the temporal dimension of the analysis—tracking stance shifts across three pandemic waves—shows that the geography of conflict evolves as events unfold, a dynamic that static snapshots of polarization cannot capture.

For geospatial policy analysis, the COVID drug discourse study provides a template for real-time monitoring of conflictual speech around contested policy interventions. The same pipeline architecture—stance detection, geographic aggregation, temporal tracking—could be applied to climate adaptation debates (where support for specific interventions varies by exposure to climate risk), vaccine mandates (where school-district-level compliance correlates with local discourse), or housing policy (where opposition rhetoric clusters in specific neighborhoods). In each case, the NLP pipeline translates unstructured text into georeferenced indicators that can be layered with demographic, environmental, and infrastructural data in the geospatial frameworks described earlier in this chapter.

### 6.3 Bridging Partisan Language Divides and the Broader Landscape of NLP for Public Discourse

The methods described in the preceding sections focus on detecting conflict and divergence. A parallel line of work addresses the challenge of bridging partisan language divides—making the implicit meanings and associations that words carry in different communities explicit and navigable. The Bridging Dictionary (Jiang, Beeferman, Brannon, Heyward, and Roy 2024), presented at the ACM Conference on Computer-Supported Cooperative Work and Social Computing (CSCW), is an AI-generated reference that catalogs how 796 politically charged terms are used distinctively by Republican and Democratic communities on social media. For each term, the dictionary provides frequency and sentiment statistics by political group, LLM-generated summaries of how the term functions in each community's discourse, and representative examples illustrating divergent usage.

The Bridging Dictionary represents a shift from measurement to intervention: rather than simply detecting polarization, it provides a resource that journalists, policymakers, and citizens can consult to understand how their language may be received across partisan lines. This work builds on a tradition in computational social science of analyzing how language encodes and reproduces political divides. Demszky et al. (2019) analyzed polarization in tweets about 21 mass shootings and found that political polarization in social media discourse is primarily driven by partisan differences in *framing* rather than topic choice, identifying framing devices such as the contrasting use of the terms "terrorist" and "crazy" that contribute to polarization. The Bridging Dictionary makes this kind of divergence navigable at the word level; Figure 9 shows the entry for "terrorists," where Fox News and MSNBC deploy the same term with starkly different referents—Islamic extremism and foreign threats on one side, domestic right-wing violence and the January 6 Capitol riot on the other—with a 78-to-22 percent usage share split that quantifies the magnitude of the framing gap. KhudaBukhsh, Sarkar, Kamlet, and Mitchell (2021) used modern machine-translation techniques to demonstrate that the left and right communities use English words so differently that translation methods designed for cross-lingual tasks can detect the divergence. For the GeoAI-for-Policy framework, a geographically enriched version of the Bridging Dictionary—one that captures not only partisan but also regional variation in word meaning—would directly support the chapter's vision of spatially situated discourse analysis.

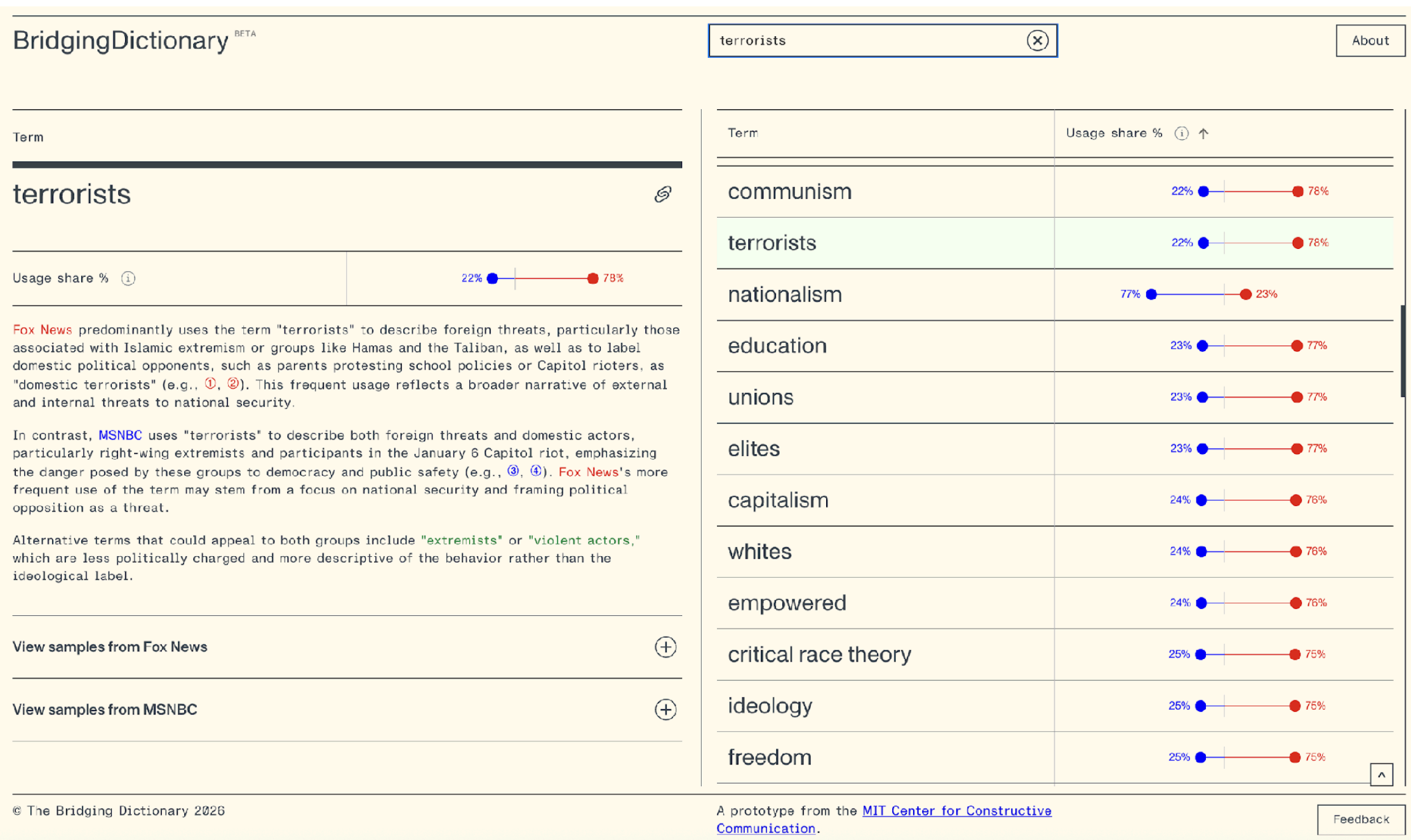

*Figure 9. The Bridging Dictionary entry for the term "terrorists," illustrating divergent partisan media framing. Fox News uses the term predominantly (78% usage share) to describe foreign threats associated with Islamic extremism and groups like Hamas and the Taliban, as well as to label domestic political opponents; MSNBC (22% usage share) applies it to both foreign threats and domestic actors, particularly right-wing extremists and participants in the January 6 Capitol riot. The right panel ranks terms by partisan usage disparity. Source: Jiang, Beeferman, Brannon, Heyward, and Roy (2024); https://dictionary.ccc-mit.org/.*

These methods sit within a broader and rapidly growing landscape of NLP for public discourse analysis. A scoping review of 154 studies on NLP and political polarization (Nemeth 2022) found that the field has been dominated by U.S. contexts (59 percent of studies), Twitter data (43 percent), and machine learning approaches (33 percent), with transformer-based models such as BERT increasingly supplanting earlier lexicon-based and bag-of-words methods. More recently, large language models such as GPT-4 have demonstrated strong zero-shot and few-shot capabilities for stance detection, sentiment analysis, and hate speech classification, reducing the annotation burden that previously limited scalable discourse analysis (Hou and Huang 2025). At the intersection of NLP and geospatial analysis, Nelson et al. (2015) developed SPoTvis, a web-based geovisual analytics tool offering a term polarity plot coupled with interactive maps that allows users to compare Twitter subthemes between states or congressional districts; Hernandez et al. (2025) introduced a geo-stance-aware recommendation framework that links geolocated social media climate discussions with real-world disaster events through spatio-temporal multilayer networks; and Hu et al. (2023) surveyed the state of the art in location reference recognition from texts, identifying geoparsing as a critical bridge between unstructured text and georeferenced analysis. The convergence of these trends—community-specific language models, scalable stance detection, geographic text analysis, and generative AI for bridging

divides—creates the methodological foundation for the spatial discourse analysis that this chapter proposes.

## 7. Future Work

The metadata underlying the knowledge graph is considerably messier than the structured representation, and enrichment of this metadata, as it is largely provided on a volunteer basis by depositors, is an area of active research. For instance, of the 43,991 geospatially tagged datasets, some datasets' metadata include precise bounding boxes while others provide little metadata beyond a single word to define their extent of their geospatial coverage (ie. some will just say "Global") in the geographic coverage field; moreover, malformed rows, inconsistent place names, and bounding boxes that do not match the stated geographic coverage are all present in the underlying records. We present these figures as proof of concept and as motivation for the place-name normalization, and we treat them as a starting point rather than a final census. The policy keyword search understates the true count by design; datasets about housing, immigration, labor markets, and environmental regulation may carry geographic metadata and policy relevance without matching our search terms, so the 7,654 figure is a lower bound (a broader search would capture more records at the cost of more false positives, and we judged the conservative approach more appropriate for establishing baseline feasibility).

The knowledge graph that we have constructed here reflects a snapshot of repository metadata at the time we built it; as such, new deposits after that date do not appear, and stale records from deleted or updated datasets may persist. Any production system built on this graph would need to ingest updates incrementally, which Dataverse's API supports but which we have not implemented. The geographic distribution of the geospatial metadata is skewed toward the United States and toward city-level data, because researchers who deposit in Harvard Dataverse are predominantly American and because city is the most commonly supplied geographic unit. The topological analysis we report will hold most firmly for American policy research; extending it to other national contexts will require supplementary geocoding, gazetteers in local languages, and administrative hierarchies from national statistics offices. We demonstrate feasibility for one national context, not global coverage.

## 8. Conclusion

The Knowledge Graph on Harvard Dataverse holds 43,991 geospatially tagged datasets, 7,654 of which address government and policy topics, all embedded in a 215,985-node network. The giant component (96.9 percent of all nodes) makes virtually any geospatial dataset reachable from any other through short chains of shared attributes. The main limitation we identify is inconsistent place names, where the same city or region appears as multiple disconnected location nodes because depositors enter different strings for the same place.

AI methods can address those inconsistencies and enrich geospatial metadata that the graph is missing. Large language models can read depositor-supplied free-text coverage strings, match ambiguous place names to standard identifiers (FIPS codes, GeoNames entries, ISO 3166 subdivisions), and extract geographic information from titles, abstracts, and keyword fields where it appears implicitly. A dataset titled "County-Level Diabetes Prevalence in Mississippi, 2010-2018" records the state, the administrative unit, and the temporal range in the title itself, even though the depositor never entered those details into the structured metadata fields. Entity-resolution models trained on gazetteers can merge the thousands of split location nodes we document, collapsing "NYC," "New York City," and "New York, NY" into a single node and creating new edges between datasets that were previously disconnected.

The contribution of this chapter is to establish the empirical foundation for that work; the Knowledge Graph on Harvard Dataverse, with 528,003 edges and 43,991 geospatially tagged datasets, provides a concrete and well-documented setting in which to develop and evaluate AI-driven metadata enrichment.

**Notes**

[1] Harvard Dataverse (https://dataverse.harvard.edu) is an open-source research data repository developed and maintained by Harvard University's Institute for Quantitative Social Science (IQSS), Harvard Library, and Harvard University Information Technology. See King, G. (2007), "An Introduction to the Dataverse Network as an Infrastructure for Data Sharing," Sociological Methods & Research 36(2), 173-199; and Crosas, M. (2011), "The Dataverse Network: An Open-Source Application for Sharing, Discovering and Preserving Data," D-Lib Magazine 17(1/2).

[2] Hogan, A. et al. (2021), "Knowledge Graphs," ACM Computing Surveys 54(4), Article 71, 1-37, doi:10.1145/3447772; Ehrlinger, L. and Wöß, W. (2016), "Towards a Definition of Knowledge Graphs," SEMANTiCS 2016 Posters and Demos, CEUR-WS Vol. 1695. On knowledge graphs for scientific data specifically, see Ammar, W. et al. (2018), "Construction of the Literature Graph in Semantic Scholar," Proceedings of NAACL-HLT 2018 (Industry Papers), 84-91; Himmelstein, D. S. et al. (2017), "Systematic Integration of Biomedical Knowledge Prioritizes Drugs for Repurposing," eLife 6, e26726, doi:10.7554/eLife.26726; and Auer, S. et al. (2018), "Towards an Open Research Knowledge Graph," doi:10.5281/zenodo.1157185.